\documentclass[runningheads]{llncs}

\usepackage{color}
\usepackage{cite}
\usepackage{amsmath}
\usepackage{amssymb}
\usepackage{booktabs} % For formal tables
\usepackage{tabu}
\usepackage{algorithm}
\usepackage{algorithmic}
\usepackage{graphicx}
\usepackage{marvosym}

\usepackage{multirow}
\usepackage{color,soul}
\usepackage[table,xcdraw]{xcolor}

\usepackage[belowskip=5pt,aboveskip=0pt]{caption}

\begin{document}
\title{AngryBERT: Joint Learning Target and Emotion for Hate Speech Detection}
% \author{Anonymous}
%
\titlerunning{AngryBERT: Joint Learning Target and Emotion for Hate Speech Detection}
% If the paper title is too long for the running head, you can set
% an abbreviated paper title here
%
\author{Md Rabiul Awal\inst{1} \and
Rui Cao\inst{2} \and
Roy Ka-Wei Lee \inst{3} \and
Sandra Mitrovi\'c\inst{4}} 

\authorrunning{M. R. Awal et al.}
% First names are abbreviated in the running head.
% If there are more than two authors, 'et al.' is used.
%
\institute{University of Saskatchewan, Saskatoon, SK, Canada \\
\email{mda219@usask.ca} \and
Singapore Management University, Singapore 188065 \\
\email{ruicao.2020@phdcs.smu.edu.sg} \and
Singapore University of Technology and Design, Singapore 487372 \\
\email{roy\_lee@sutd.edu.sg} \and
Dalle Molle Institute for Artificial Intelligence, Lugano, Switzerland \\ \email{sandra.mitrovic@idsia.ch}}

\maketitle              % typeset the header of the contribution
\begin{abstract}

Automated hate speech detection in social media is a challenging task that has recently gained significant traction in the data mining and Natural Language Processing community. However, most of the existing methods adopt a supervised approach that depended heavily on the annotated hate speech datasets, which are imbalanced and often lack training samples for hateful content. This paper addresses the research gaps by proposing a novel multitask learning-based model, \textsf{AngryBERT}, which jointly learns hate speech detection with sentiment classification and target identification as secondary relevant tasks. We conduct extensive experiments to augment three commonly-used hate speech detection datasets. Our experiment results show that \textsf{AngryBERT} outperforms state-of-the-art single-task-learning and multitask learning baselines. We conduct ablation studies and case studies to empirically examine the strengths and characteristics of our \textsf{AngryBERT} model and show that the secondary tasks are able to improve hate speech detection. 

\end{abstract}

\section{Introduction}
\textbf{Motivation.} The sharp increase in online hate speeches has raised concerns globally as the spread of such toxic content and misbehavior have not only sowed discord among individuals or communities online but also resulted in violent hate crimes. Therefore, it is a pressing issue to detect and curb hate speech in online social media. Researchers have proposed many traditional and deep learning hate speech classification methods to detect hate speeches in online social media automatically \cite{fortuna2018survey}. Specifically, the existing deep learning methods have achieved promising performance in the hate speech detection task \cite{cao2020deephate}. However, most of these supervised methods depended heavily on the annotated hate speech datasets, which are imbalanced and often lack training samples for hateful content \cite{arango2019hate}. A potential solution to address the challenges of imbalanced datasets is to perform data augmentation for the class with fewer training samples \cite{cao2020hategan}. Nevertheless, the existing data augmentation methods have shown limited improvement in hate speech detection.

\textbf{Research Objectives.} In this paper, we adopt a different approach to address the research gaps. We propose a novel multitask learning-based model, \textsf{AngryBERT} \footnote{code implementation: https://gitlab.com/bottle\_shop/safe/angrybert}, which jointly learns hate speech detection with secondary relevant tasks. Multitask learning (MTL) \cite{zhang2017survey} is a machine learning paradigm that aims to leverage useful information in multiple related tasks to help improve the generalization performance of all the tasks. Earlier studies have shown that MTL improved the performance of text classification tasks even when training with inadequate samples \cite{zhang2017survey}. Similarly, the intuition of our \textsf{AngryBERT} model is that the auxiliary datasets from the secondary relevant tasks supplement the limited hateful samples of the datasets used for the main hate speech detection task. Specifically, we utilize emotion classification \cite{mohammad2018semeval} and hateful target identification \cite{ElSherief2018HateLA,zampieri2019semeval} as the secondary tasks in our proposed model. Emotion classification is a relevant task as previous studies have demonstrated that sentiments are useful features in hate speech classification \cite{fortuna2018survey,cao2020deephate}. Hateful target identification is an extension to the hate speech detection task where it aims to identify the target group or individual victim of the hateful content. Another key component in our \textsf{AngryBERT} model is the BERT transformer model \cite{devlin2019bert}, which is fine-tuned and used as the layer to share knowledge across various tasks. To the best of our knowledge, \textsf{AngryBERT} is the first model that uses a pre-trained and fine-tuned language model in a MTL framework for hate speech detection.

\textbf{Contributions.} We summarize our paper contribution as follows: (i) We propose a novel MTL and BERT-based model call \textsf{AngryBERT}, which jointly learns hate speech detection with secondary relevant tasks. (ii) We conduct extensive experiments on three commonly-used hate speech detection datasets. Our experiment results show that \textsf{AngryBERT} outperforms the state-of-the-art single-task and multitask baselines in hate speech detection. (iii) We identify case studies to demonstrate that \textsf{AngryBERT} is able to detect hate speeches accurately and identify the target of the hate speech and the emotion expressed. This showcases \textsf{AngryBERT}'s potential to provide some form of explainability to the hate speech detection task.

\section{Related Work}
In this section, we reviewed two groups of literature relevant to our study, namely, (i) existing studies on automated hate speech detection and (ii) multitask learning (MTL) for natural language processing (NLP) tasks.

Automatic detection of hate speech has received considerable attention from data mining, information retrieval, and NLP research communities. Earlier works have explored hand-crafted and canonical NLP features for automatic hate detection \cite{fortuna2018survey,DavidsonWMW17, waseem2016hateful, waseem2016you}. In recent years, researchers have proposed deep learning methods to extract latent features more effectively for hate speech detection \cite{fortuna2018survey, ParkF17,zhang2018detecting,badjatiya2017deep}. Most of these methods adopt a supervised approach that heavily depends on labeled datasets for training, which is a challenge as existing hate speech datasets are highly imbalanced and lack training examples for hateful content.

%More recent works have proved that the latent patterns can be much more effectively encoded with the deep architectures \cite{mehdad2016characters}. Consequently, RNNs \cite{mehdad2016characters, FountaCKBVL19}, CNNs \cite{gamback2017using, ParkF17}, LSTM \cite{badjatiya2017deep, grondahl2018all}, CNN+GRU  \cite{zhang2018detecting, grondahl2018all} have been exploited for proposing different automatic hate speech detection methods. 

%On the other hand, multi-task learning based on neural architecture has seen immense expansion lately 
% \cite{luong2015multi,liu2016recurrent,zhang2017generalized,xiao2018gated,xiao2018learning,waseem2018bridging,tang2019multi,liu2019multi,rajamanickam2020joint}. 
%for solving different NLP problems, 

MTL is a popular machine learning paradigm that has been explored and applied in various NLP problems, such as text classification \cite{liu2016recurrent, liu2019multi}, etc. MTL has also been applied to abusive speech detection \cite{rajamanickam2020joint,waseem2018bridging}. Waseem et al. \cite{waseem2018bridging} proposed a fully-shared MTL model, which all tasks utilize the same fully shared features, to performed hate speech detection on three hate speech datasets. Unlike \cite{waseem2018bridging}, our proposed \textsf{AngryBERT} model adopts the shared-private scheme, which model distinguishes between task-dependent and task-invariant (shared) features to perform the primary and secondary tasks. Furthermore, unlike \cite{waseem2018bridging} that only considered hate speech detection task and datasets, our proposed model used other relevant auxiliary tasks and dataset to improve the primary hate speech detection task. Closer to our study, Rajamanickam et al. \cite{rajamanickam2020joint} proposed a shared-private MTL framework that utilized a stacked BiLSTM encoder as the shared layer and attention mechanism for intra-task learning. The framework is trained on a hate speech detection dataset for the primary task and emotion detection as the secondary relevant task. Different from \cite{rajamanickam2020joint}, our \textsf{AngryBERT} model adopted BERT \cite{devlin2019bert} as the shared layer, and is trained on both emotion classification and hateful target identification as secondary tasks to aid hate speech detection.

\section{Datasets and Tasks}
Previous studies have shown that the relevance of tasks in an MTL framework affects the model's stability of training and performance \cite{zhang2017survey}. According to the definition of hate speech, there are two main characteristics of hate speech: (i) offensive language that (ii) targets individuals or groups. Considering the two aspects, we select two secondary tasks relevant to hate speech detection: emotion classification and target identification. Offensive language usually involves negative sentiments. Therefore, emotions in tweets can serve as complementary information for hate speech detection~\cite{cao2020deephate}. Our goal is to train a network that can extract emotions hidden in tweets using the emotion classification task. For the target identification task, we aim to train the model to identify targets in a text. Co-trained with these two secondary tasks, MTL models will be capable of extracting emotions and target groups or individuals in tweets, which facilitates hate speech detection indirectly. In the remaining parts of this section, we discuss the datasets involved to train the \textsf{AngryBERT} model and MTL baselines. Table \ref{tab:dataset} shows the statistical summary of the datasets.

\subsection{Primary Task and Datasets}
The primary task of \textsf{AngryBERT} is hate speech detection. Therefore, we train and evaluate our proposed model on three publicly available hateful and abusive speech datasets,  namely, WZ-LS~\cite{ParkF17}, DT~\cite{DavidsonWMW17}, and FOUNTA~\cite{FountaDCLBSVSK18}.

{\bf WZ-LS}~\cite{ParkF17}: Park et al.~\cite{ParkF17} combined two Twitter datasets \cite{waseem2016hateful,waseem2016you} to form the WZ-LS dataset. We retrieve the tweets' text using Twitter's APIs and the tweet ids release in \cite{ParkF17}. However, some of the tweets have been deleted by Twitter due to their inappropriate content. Thus, our dataset is slightly smaller than the original dataset reported in \cite{ParkF17}.

{\bf DT}~\cite{DavidsonWMW17}: Davidson et al.~\cite{DavidsonWMW17} The researchers constructed the DT Twitter dataset, which manually labeled and categorized tweets into three categories: offensive, hate, and neither. %argued that hate speech should be differentiated from offensive tweets; some tweets may contain abusive language without targeting at a specific group or individual.

{\bf FOUNTA}~\cite{FountaDCLBSVSK18}: The FOUNTA dataset is a human-annotated dataset that went through two rounds of annotations. Awal et al.~\cite{awal2020analyzing} found that there were duplicated tweets in FOUNTA dataset as the dataset annotators have included retweets in their dataset. For our experiments, we remove the retweets resulting in the distribution in Table \ref{tab:dataset}.

\begin{table}[t]
\centering
  \caption{Statistic information about datasets in experiments}
  \label{tab:dataset}
  %\resizebox{1\columnwidth}{!}{
  \begin{tabular}{p{6em}|c|p{26em}}
    %\toprule
    \hline
    \textbf{Dataset} & \textbf{\#tweets} & \textbf{Classes (\#tweets)}\\\hline\hline
    %\midrule
    %\midrule
    DT & 24,783 & hate(1,430), offensive(19,190), neither(4,163)\\\hline
    WZ-LS & 16,035 & racism(1923), sexism(3,079), neither(11,033)\\\hline

    FOUNTA & 89,990 & normal (53,011), abusive (19,232), spam (13,840), hate (3,907)\\\hline
    %\midrule
    HateLingo &  5,680 & disability(257), ethnicity(351), gender(2841), religion(1590), sexual orientation(641) \\\hline
    SemEval\_A &  10,983 & anger(2544), anticipation(978), disgust(2602), fear(1242), joy(2477), love(700), optimism(1984), pessimism(795), sadness(2008), surprise(361), trust(357) \\\hline
    %\midrule
    OffensEval\_C & 4,089 & individual(2,507), group(1,152), other(430)\\\hline
    %\midrule
  %\bottomrule
\end{tabular}%}
\end{table}

\subsection{Secondary Tasks and Datasets.}
Three publicly available Twitter datasets are selected for the secondary tasks: SemEval\_A~\cite{mohammad2018semeval}, HateLingo~\cite{ElSherief2018HateLA}, and OffensEval\_C~\cite{zampieri2019semeval}. 
%emotion classification and target identification 

\textbf{SemEval\_A}~\cite{mohammad2018semeval}: Mohammad et al. collected and annotated a Twitter dataset that supported array of subtasks on inferring the affectual state of a person from
their tweet. We perform emotion classification task using this Twitter dataset.

\textbf{HateLingo}~\cite{ElSherief2018HateLA}: ElSherief et al. collected the HateLingo dataset that identifies the target of hate speeches. We perform hate speech target group identification task using the HateLingo dataset. Specifically, the task aims to identify the target group in a given hateful tweet. 

\textbf{OffensEval\_C}~\cite{zampieri2019semeval}: Zampieri et al. proposed the OffenEval\_C dataset, which categorize the targets of abusive tweets into \textit{individual}, \textit{group}, or \textit{other}. Similarly, our proposed model is trained on this dataset for target identification task.

\vspace{-1em}

\section{Proposed Model}
\subsection{Problem Formulation}
Essentially, hate speech detection (i.e., primary task) and the relevant secondary tasks can be generalized as text classification tasks. Therefore, we define a general problem formulation of text classification tasks under the MTL setting. Assume we have $K$ tasks and the input for the i-th task is: $S_i=\{s^i_1,s^i_2\dots, s^i_n\}, i\in\{1,2,\dots,K\}$, where $n$ is the length of the sentence. For the i-th task, the goal is to correctly classify the input text into: $C=\{c^i_1,c^i_2\dots,c^i_m\}$, where m is the number of classes of task $i$. 

\subsection{Architecture of AngryBERT}
\begin{figure}[t]
    \centering
	\includegraphics[width=1 \textwidth]{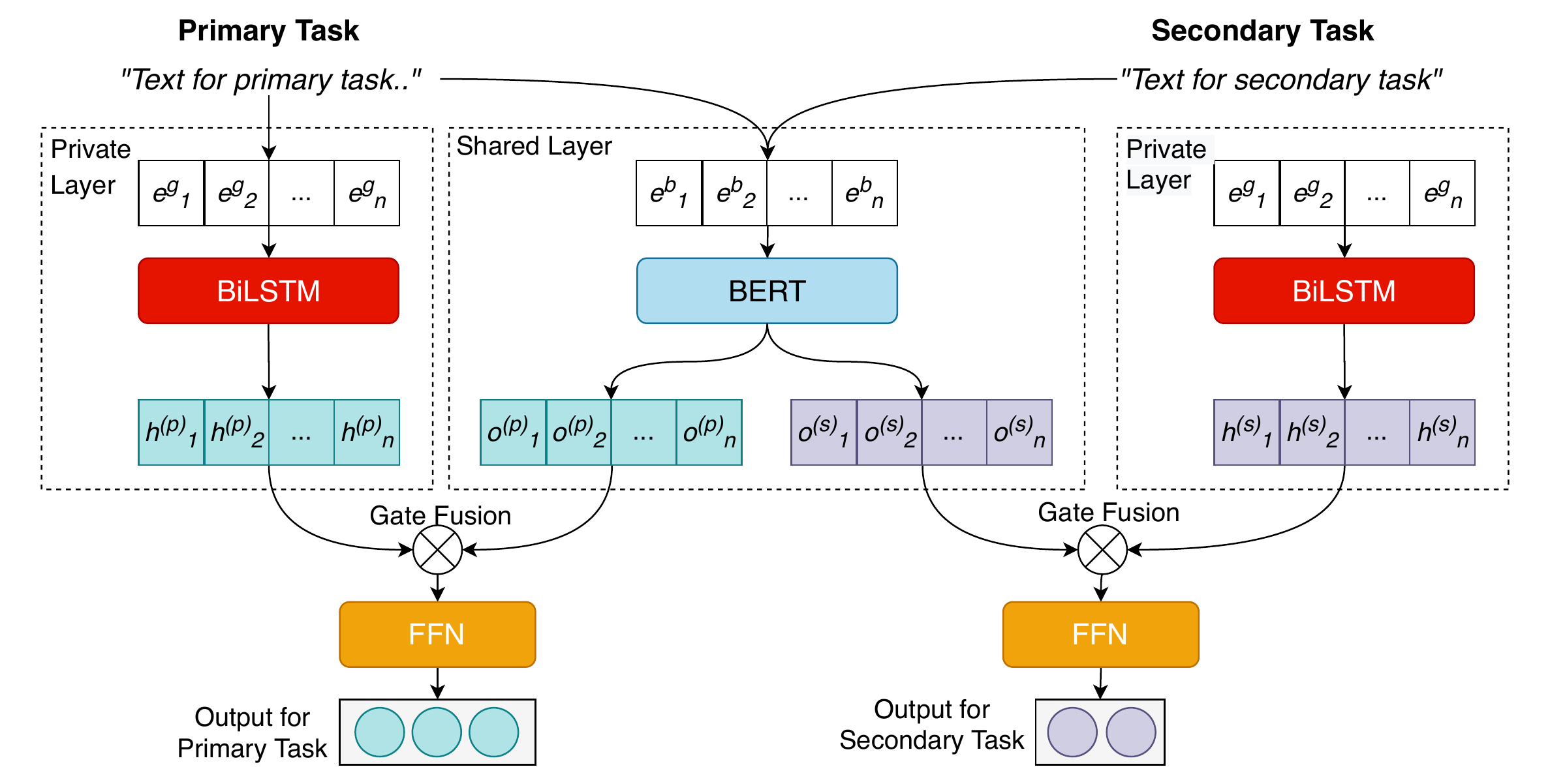}
	\caption{The overall architecture of \textsf{AngryBERT} model}
	\label{fig:framework_new}
\end{figure}

Figure~\ref{fig:framework_new} illustrates the overall architecture of \textsf{AngryBERT} model. We adopt a shared-private MTL setting where a shared layer is encouraged to learn the task-invariant features while the private layers aim to learn the task-specific representations. The gate fusion mechanism aggregates the shared and private information. Finally, the joint representation of each task is fed into their classification layer, respectively. To simplify our discussion, we ignore the superscript for each task in the rest of this section.

%In this part, we discuss about the architecture of the proposed AngryBert. It consists of one shared layer, various private layers and classification layers for each task. Shared layer is encouraged to learn the task-invariant features while the private layers aim to learn the task-specific representations. The shared and private information are aggregated by gate fusion mechanism. Finally the joint representation of each task is fed into the their classification layer respectively. To be more general, we ignore the superscript for each task in this part.

\textbf{Shared Layer} Here we exploit the pre-trained BERT model as the shared layer. Given a sentence, it is first tokenized using the default tokenizer of BERT then transformed into pre-trained BERT embeddings: $E_B=\{e^b_1,e^b_2,\dots,e^b_n\}$, $e^b_i\in R^{768}$. These embeddings are sent to a pre-trained BERT model. We use the output from the \textsf{[CLS]} token as the representation from the shared layer, denoted as $o_1\in R^d$:
    \begin{equation}
        o_1=BERT(E_B)
    \end{equation}
    
\textbf{Private Layer} For each task, a private layer is used to learn the task-specific representation. In order to fully exploit contexts of each word, a Bi-directional Long-Short Term Memory Network (Bi-LSTM) is applied. Each word of the sentence is first embedded using GloVe Embedding: $E_G=\{e^g_1,e^g_2$, $\dots,e^g_n\}$, $e^g_i\in R^{300}$. The embeddings are sent to the Bi-LSTM to learn the sequential information. The concatenation of final hidden states from forward and backward path is used as the latent representation learnt from the private layer, denoted as $h_n\in R^d$:
    \begin{equation}
        h_n=Bi-LSTM(E_G)
    \end{equation}
    
\textbf{Gate Fusion} After learning the respective representations from shared and private layers, we exploit the gate mechanism for feature fusion. Instead of directly assign a weight for each vector, the gate fusion mechanism allows each position of vectors to have different contribution to the prediction. The joint representation from gate fusion is computed as below:
    \begin{equation}
        \alpha=\sigma(W_Lo_1+W_Bh_n+b_g)
    \end{equation}
    \begin{equation}
        J=\alpha L+(1-\alpha)B
    \end{equation}
    where $W_L$, $W_B$ and $b_g$ are parameters to be learnt. $\alpha \in R^{d}$, which is of the same dimension as $h_n$ and $o_1$, is the attention vector. It controls the proportion of information from the private and shared flow.
    
\textbf{Classification Layer} For each task, we feed the joint representation after information aggregation to its classification layer. The classification layer is a Multi-Layer Perceptron (MLP) follow by a Softmax layer for normalization. 
    \begin{equation}
        M=ReLU(W_f J+b_f)
    \end{equation}
    \begin{equation}
        O=softmax(W_e M +b_e)
    \end{equation}
    where $W_f$, $W_e$ and $b_f$, $b_e$ are weights and biases to be learnt. The final prediction is $O\in R^{m}$ and each position of $O$ denotes the confidence score for each class. Non linear activation function and weight normalization are used between two linear projection layers. Dropout is applied in order to avoid overfitting in the classification layers.

\subsection{Training of AngryBERT}
In this part, we describe the loss function of individual tasks and the training for \textsf{AngryBERT} under the MTL setting.

\textbf{Single Task Loss} For each task, cross entropy is used as the loss function. The loss of the i-th task is:
    \begin{equation}
        M_i=\sum_{t=1}^{N_i}Cross-Entropy(O_t^i,\hat{O^i_t})
    \end{equation}
    where $\hat{O^i_t}$ is the ground-truth class for the t-th instance of task $i$ and $N_i$ is the number of training instances for the i-th task. For all tasks, we obtain: $M=\{M_1,M_2,\dots,M_K\}$.

\textbf{Multi-task Loss} There are several objectives involved: the primary task and secondary tasks. Rather than averaging all losses, we consider different speeds of divergence of tasks. The objective function is a weighted average of losses from different tasks: 
    \begin{equation}
        \Phi =\sum_{i=1}^{K} \beta_i M_i
    \end{equation}
    where weights $\beta_i$ are learnt end-to-end, which represents the contribution from task $i$ to the multitask loss. By exploiting multitask loss, tasks have different importance for parameter updating, which mitigates the issue of different speeds of convergence. All tasks are trained with the same number of epochs.

\section{Experiments}
In this section, we will first describe the settings of experiments conducted to evaluate our \textsf{AngryBERT} model. Next, we discuss the experiment results and assess how \textsf{AngryBERT} fares against other state-of-the-art baselines. We conduct more in-depth ablation studies on the various tasks co-trained with the primary hate speech detection task in the \textsf{AngryBERT}. We demonstrate interesting case studies where the tweets' predicted labels for various tasks co-trained in \textsf{AngryBERT} presented.

\subsection{Baselines.} We compare \textsf{AngryBERT} with the state-of-the-art hate speech classification baselines and multitask learning text classification models:

\begin{itemize}
    \item \textbf{CNN}: Previous studies have utilized CNN to achieve good performance in hate speech detection \cite{badjatiya2017deep}. We train a CNN model with word embeddings as input.  
    \item \textbf{LSTM}: The LSTM model, is another model that was commonly explored in previous hate speech detection studies~\cite{badjatiya2017deep}. Similarly, we train a LSTM model with word embeddings as input.
    \item \textbf{HybridCNN}: We replicate the HybridCNN model proposed by Park and Fung~\cite{ParkF17} for comparison. The HybridCNN model trains CNN over both word and character embeddings for hate speech detection.
    \item \textbf{CNN-GRU}: The CNN-GRU model was proposed in a recent study by Zhang et al.~\cite{zhang2018detecting} is also replicated in our study as a baseline. The CNN-GRU model takes word embeddings as input.
    \item \textbf{DeepHate}: The DeepHate model was proposed in a recent study by Cao et al.~\cite{cao2020deephate}. The DeepHate model trains on semantics, sentiment, and topical features for hate speech detection.
    \item \textbf{BERT}: BERT~\cite{devlin2019bert} is a contextualized word representation model that is based on a masked language model and pre-trained using bidirectional transformers. For our study, we fine-tune the pre-trained BERT model using the train set and subsequently perform classification on tweets in the test set.
    \item \textbf{SP-MTL}: Liu et al.~\cite{liu2016recurrent} proposed the SP-MTL model, which is a Recurrent Neural Network (RNN) based multitask learning model for text classification tasks. We trained the SP-MTL model with the same tasks as our \textsf{AngryBERT} model.
    \item \textbf{MT-DNN}: Liu et al.~\cite{liu2019multi} proposed the  Multi-Task Deep Neural Network (MT-DNN), which combined multitask learning and language model pre-training for language representation learning. We replicated the MT-DNN as a baseline in our study. Similarly, we trained the MT-DNN with the same tasks as our \textsf{AngryBERT} model.
    \item \textbf{MTL-GatedDEncoder}: Rajamanickam et al.~\cite{rajamanickam2020joint} proposed a shared-private MTL framework that utilized a stacked BiLSTM encoder as the shared layer and attention mechanism for hate speech detection. This is the state-of-the-art MTL baseline for hate speech detection.
\end{itemize}

\subsection{Evaluation Metrics.} Similar to most existing hate speech detection studies, we use micro averaging precision, recall, and F1 score as the evaluation metrics. Five-fold cross-validation is used in our experiments, and the average results are reported. %Micro averaging is preferred in our experiments as hate speech datasets are imbalanced. 

\begin{table}[t]
\centering
  \caption{Experiment results of AngryBERT and baselines on DT, WZ-LS, and FOUNTA datasets. ``\#'' denotes MTL models that co-trained with other secondary tasks.}
  \label{tab:main_results}
\begin{tabular}{p{3.6cm} | ccc|ccc|ccc}
    \hline
    & \multicolumn{3}{c|}{\cellcolor[HTML]{B0E3E6}\textbf{DT}} & \multicolumn{3}{c|}{\cellcolor[HTML]{B1DDF0}\textbf{WZ-LS}} & \multicolumn{3}{c}{\cellcolor[HTML]{D0CEE2}\textbf{FOUNTA}} \\
    \textbf{Model} & \cellcolor[HTML]{B0E3E6}\textbf{Prec} & 
    \cellcolor[HTML]{B0E3E6}\textbf{Rec} & 
    \cellcolor[HTML]{B0E3E6}\textbf{F1} & 
    \cellcolor[HTML]{B1DDF0}\textbf{Prec} & 
    \cellcolor[HTML]{B1DDF0}\textbf{Rec} & 
    \cellcolor[HTML]{B1DDF0}\textbf{F1}  &
    \cellcolor[HTML]{D0CEE2}\textbf{Prec} & 
    \cellcolor[HTML]{D0CEE2}\textbf{Rec} & 
    \cellcolor[HTML]{D0CEE2}\textbf{F1}  \\
    \hline\hline
    CNN & 89.32 & 90.07 & 89.35 & 80.63 & 78.35 & 78.21 & 79.97 & 80.35 & 79.84 \\
    %CNN-C & 78.94 & 81.31 & 78.77  & 73.44 & 74.00 & 71.85 & 70.69 & 71.38 & 67.57\\ 
    %CNN-B & 59.96 & 77.43 & 67.58 & 47.34 & 68.81 & 56.09 & 53.83 & 59.60 & 52.83 \\ 
    LSTM & 89.58 & 90.26 & 89.56 & 80.43 & 77.54 & 77.27 & 80.24 & 81.18 & 80.22 \\ 
    %LSTM-C & 77.37 & 79.50 & 76.58 & 72.87 & 74.95 & 69.36 & 70.13 & 70.47 & 65.47 \\ 
    %LSTM-B & 59.96 & 77.43 & 67.58 & 47.34 & 68.81 & 56.09 & 51.32 & 59.76 & 52.25 \\ 
    HybridCNN & 88.65 & 89.91 & 88.85 & 80.71 & 78.91 & 78.3 & 79.86 & 80.52 & 79.86 \\ 
    CNN-GRU & 88.89 & 89.80 & 88.91 & 80.85 & 77.05 & 77.12 & 79.96 & 80.73 & 79.99 \\ 
    DeepHate & 89.97 & 90.39 & 89.92 & 77.95 & 79.48 & 78.19 & 78.95 & 80.43 & 79.09 \\ 
    BERT & 90.35 & 90.53 & 90.34 & \textbf{83.25} & 80.05 & 79.95 & 79.69 & 80.03 & 79.79 \\ \hline
    %\multirow{2}{*}{DT+HateLingo+Emotion+SemEval\_C} 
    SP-MTL\# & 89.44 & 90.22 & 89.44 &  81.11 & \textbf{81.59} & 80.68 & 80.46 & 81.65 & 80.66\\
    MT-DNN\# & 90.29 & 90.69 & 90.31 & 83.05 & 80.25 & 80.18 & 80.66 & 81.64 & 80.72 \\
    MTL-GatedDEncoder\# & 89.20 & 89.55 & 89.22 & 81.33 & 78.62 & 78.18 & 80.00 & 81.33 & 80.08\\
    AngryBERT\# & \textbf{90.71} & \textbf{91.14} & \textbf{90.71} & 83.19 & 81.45 & \textbf{81.25} & \textbf{81.00} & \textbf{81.82} & \textbf{81.08}  \\
    \hline
\end{tabular}
\end{table}

\subsection{Experiment Results.} Table \ref{tab:main_results} shows the experiment results on DT, WZ-LS, and FOUNTA datasets. In the table, the highest figures are highlighted in \textbf{bold}. We observe that \textsf{AngryBERT} outperformed the state-of-the-art single and multitask baselines in Micro-F1 scores. We observed that the single task BERT model is able to achieve good performance in hate speech detection, outperforming the other single task baselines for DT and WZ-LS datasets. Nevertheless, \textsf{AngryBERT} outperformed the BERT baseline by leveraging the BERT language model to learn shared knowledge across tasks.

Comparing the single-task baselines with the MTL-based models, we noted that the MTL-based models are able to outperform most single-task baselines across the three hate speech datasets. The observation shows the advantage to co-train the hate speech detection task with other secondary tasks in a multitask setting. AngryBERT is observed to outperform the state-of-the-art MTL hate speech detection model, MTL-GatedDEncoder, and other MTL text classification models. The good performance demonstrates BERT's strength as the shared layer in the multitask learning architecture. 

It is worth noting that there are differences between HybridCNN and CNN-GRU models in our experiments and the results reported in previous studies \cite{ParkF17,zhang2018detecting}. For instance, earlier studies for HybridCNN \cite{ParkF17} and CNN-GRU \cite{zhang2018detecting} had conducted experiments on the WZ-LS dataset. However, we did not cite the previous scores directly as some of the tweets in WZ-LS have been deleted. Similarly, CNN-GRU was also previously applied to the DT dataset. However, in the previous work \cite{zhang2018detecting}, the researchers have cast the problem into binary classification by re-labeling the offensive tweets as non-hate. In our experiment, we perform the classification based on the original DT dataset \cite{DavidsonWMW17}. Therefore, we replicated the HybridCNN and CNN-GRU models and applied them to the updated WZ-LS dataset and original DT dataset.  

\subsection{Ablation Study.}

The \textsf{AngryBERT} model is co-trained with several secondary tasks. In this evaluation, we perform an ablation study to investigate the effects of co-training the hate speech detection tasks with different secondary tasks.

\begin{figure}[t]
    \centering
	\includegraphics[scale=0.5]{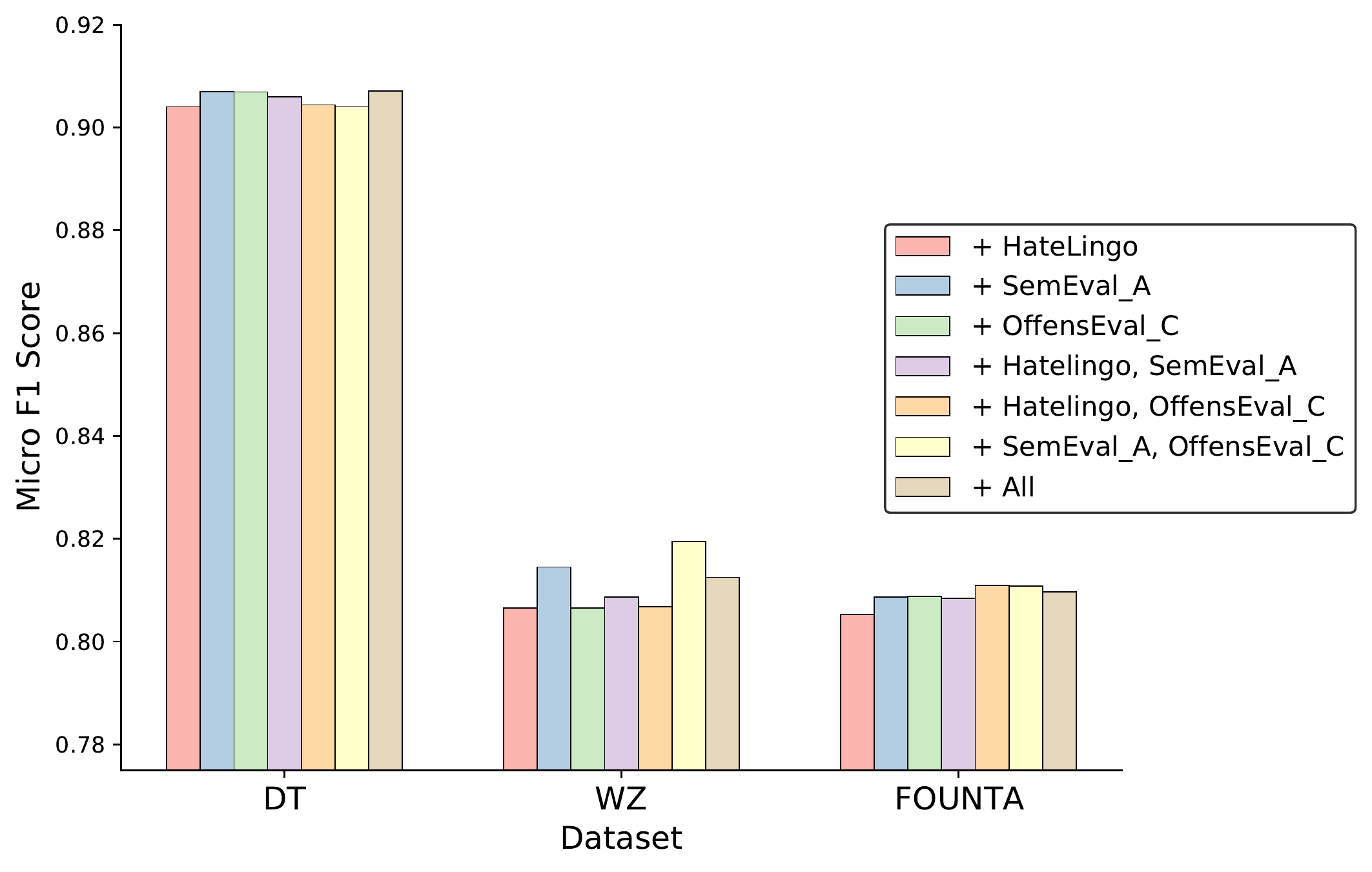}
	\caption{Micro-F1 scores of \textsf{AngryBERT} model for various hate speech datasets co-trained with various combinations of secondary task datasets}
	\label{fig:ablation}
\end{figure}

Figure~\ref{fig:ablation} shows the Micro-F1 scores of the \textsf{AngryBERT} model for different hate speech datasets co-trained with various secondary tasks. For example, the red bars show the \textsf{AngryBERT} co-training on hate speech detection tasks and the target identification task using the HateLingo dataset. We noted that the different hate speech datasets would require different task combination to achieve the best hate speech detection results. For instance, in the DT dataset, co-training the DT dataset with either the target identification task using OffensEval\_C or SemEval\_A will achieve similar performance as co-training all secondary tasks. For the FOUNTA dataset, co-training with the combinations of HateLingo + OffensEval\_C or SemEval\_A + OffensEval\_C will achieve the best performance. The WZ-LS dataset's best performance is achieved by co-training with the SemEval\_A + OffensEval\_C, and co-training with only SemEval\_A outperforms co-training with all secondary task datasets. Nevertheless, co-training with any combinations of the secondary tasks in \textsf{AngryBERT} outperforms the single-task methods in hate speech detection. These observations highlighted the intricacy of task selections for performing hate speech detection in a MTL setting. For future works, we will explore developing better approaches to automatically select the optimal combination of co-training tasks for hate speech detection.   

\subsection{Case Studies}
To better understand how the secondary tasks could help in the hate speech detection task, we qualitatively examine some sample predictions of the \textbf{AngryBERT} model. Table~\ref{tab:case_dt},\ref{tab:case_wz}, and \ref{tab:case_founta} shows example posts from DT, WZ-LS, and FOUNTA datasets respectively. In each example post, we display the actual label and predicted label from \textbf{AngryBERT} model. The correct predictions are marked in green font, while the incorrect predictions are represented in red font. Besides the hate speech predictions, we also display the predicted labels of various secondary tasks. Specifically, we highlighted the keywords in the given post that might have influenced the predicted target in HateLingo dataset.

\begin{table}[t]
\caption{Samples of AngryBERT predictions on DT dataset}
\label{tab:case_dt}
  \resizebox{\textwidth}{!}{
  \begin{tabular}{p{15em}ccccc}
    \hline
    \textbf{\cellcolor[HTML]{B0E3E6}Tweet} & \textbf{\cellcolor[HTML]{B0E3E6}DT} & \textbf{\cellcolor[HTML]{B0E3E6}DT} & \textbf{\cellcolor[HTML]{B0E3E6}SemEval\_A} & \textbf{\cellcolor[HTML]{B0E3E6}HateLingo}  & \textbf{\cellcolor[HTML]{B0E3E6}OffenEval\_C} \\
     \cellcolor[HTML]{B0E3E6} & \textbf{(\cellcolor[HTML]{B0E3E6}Actual)} & \textbf{\cellcolor[HTML]{B0E3E6}(Predict)} & \textbf{\cellcolor[HTML]{B0E3E6}(Predict)} & \textbf{\cellcolor[HTML]{B0E3E6}(Predict)} & \textbf{\cellcolor[HTML]{B0E3E6}(Predict)} \\
    \hline \hline
    [\textit{USER}] f*ck outta here and go put some more trash a*s ink on your \hl{faggot} a*s self p*ssy & hateful & \color[HTML]{009901} hateful & anger, disgust & \hl{sexual orientation}  & individual \\ \hline
    RT [\textit{USER}] We \hl{Muslims} have no military honour whatsoever we are sub human savages that slaughter unarmed men women and children & hateful & \color[HTML]{009901} hateful & anger, disgust & \hl{religion}  & group \\ \hline
    RT [\textit{USER}] I hate these Mone Davis commercials B*tch is gonna end up either a dyke or a loser like every other \hl{female} & offensive & \color[HTML]{FE0000} hateful & anger, disgust & \hl{gender} & individual \\ \hline
\end{tabular}
}
\end{table}

\begin{table}[t]
\caption{Samples of AngryBERT predictions on WZ-LS dataset}
\label{tab:case_wz}
  \resizebox{\textwidth}{!}{
  \begin{tabular}{p{15em}ccccc}
    \hline
    \textbf{\cellcolor[HTML]{B1DDF0}Tweet} & \textbf{\cellcolor[HTML]{B1DDF0}WZ-LS} & \textbf{\cellcolor[HTML]{B1DDF0}WZ-LS} & \textbf{\cellcolor[HTML]{B1DDF0}SemEval\_A} & \textbf{\cellcolor[HTML]{B1DDF0}HateLingo}  & \textbf{\cellcolor[HTML]{B1DDF0}OffenEval\_C} \\
     \cellcolor[HTML]{B1DDF0} & \textbf{(\cellcolor[HTML]{B1DDF0}Actual)} & \textbf{\cellcolor[HTML]{B1DDF0}(Predict)} & \textbf{\cellcolor[HTML]{B1DDF0}(Predict)} & \textbf{\cellcolor[HTML]{B1DDF0}(Predict)} & \textbf{\cellcolor[HTML]{B1DDF0}(Predict)} \\
    \hline \hline
    [\textit{USER}] Of course \hl{Muslim} \hl{religious} bigots like you think that is okay & racism & \color[HTML]{009901} racism & disgust, fear & \hl{religion}  & group \\ \hline
    [\textit{USER}] And if you are going to follow a \hl{prophet} that approved of collateral damage then do not complain about collateral damage & racism & \color[HTML]{009901} racism & anger, disgust, fear & \hl{religion} & individual \\\hline
    [\textit{USER}] a lying taquiyya \hl{b*tch} with zero followers opened an account to feed me bullshit & sexism & \color[HTML]{FE0000} neither & disgust, anger & \hl{gender} & individual  \\ \hline
    
\end{tabular}
}
\end{table}

\begin{table}[t]
\caption{Samples of AngryBERT predictions on FOUNTA datasets.}
\label{tab:case_founta}
  \resizebox{\textwidth}{!}{
  \begin{tabular}{p{15em}ccccc}
    \hline
    \textbf{\cellcolor[HTML]{D0CEE2}Tweet} & \textbf{\cellcolor[HTML]{D0CEE2}FOUNTA} & \textbf{\cellcolor[HTML]{D0CEE2}FOUNTA} & \textbf{\cellcolor[HTML]{D0CEE2}SemEval\_A} & \textbf{\cellcolor[HTML]{D0CEE2}HateLingo}  & \textbf{\cellcolor[HTML]{D0CEE2}OffenEval\_C} \\
     \cellcolor[HTML]{D0CEE2} & \textbf{(\cellcolor[HTML]{D0CEE2}Actual)} & \textbf{\cellcolor[HTML]{D0CEE2}(Predict)} & \textbf{\cellcolor[HTML]{D0CEE2}(Predict)} & \textbf{\cellcolor[HTML]{D0CEE2}(Predict)} & \textbf{\cellcolor[HTML]{D0CEE2}(Predict)} \\
    \hline \hline
    
     RT [\textit{USER}] I hope a tornado destroys your house you f*cking \hl{Jew} &  hateful & \color[HTML]{009901}hateful & anger, disgust & \hl{ethnicity}  & individual \\\hline
     [\textit{USER}] and added to his discomfort he is mightily pissed off at having to pay \hl{tampon} tax the cheek & normal & \color[HTML]{009901} normal & anger, disgust & \hl{gender} & individual \\\hline
     RT [\textit{USER}] They are F*CKING EVIL I DESPISE \hl{liberals} They KILL RAPE ASSAULT & abusive & \color[HTML]{FE0000} hateful & anger, disgust & \hl{gender} & group \\\hline
\end{tabular}
}
\end{table}

From the example posts, we observed that the secondary tasks profoundly impact \textsf{AngryBERT}'s hate speech detection performance. For instance, we noted that most of the predicted hateful posts are also predicted to contain ``anger'' and ``disgust'' emotions using the secondary task emotion classifier co-trained using the SemEval\_A dataset. We postulate that the emotions captured by the \textsf{AngryBERT} model have helped the model in identifying hateful content as the two emotions are commonly exhibited in online hate speeches and abusive tweets. Another interesting observation is the identification of targets in hate speeches. We observe that the secondary task of target identification classifier co-trained using HateLingo dataset is able to predict the target in a hate speech reasonably. For example, the second tweet in Table~\ref{tab:case_dt} is a hateful tweet against Muslims, and the target identification classifier predicted ``religion'' as the target in this tweet. Although the  \textsf{AngryBERT} has outperformed the state-of-the-art baselines in hate speech predictions, the model also made some incorrect predictions. However, we noted that as the ground truth labels of the incorrect predictions look contestable. For example, the last tweet in Table~\ref{tab:case_founta} seems hateful, but it was instead annotated as abusive. 

The interesting predictions from secondary relevant tasks seems to provide a form of explanation that could help us understand the context when a tweet is predicted to be hateful. For future work, we will explore building explainable models that utilize the prediction of secondary tasks as supplementary information to aid explaining hate speech detection.

\section{Conclusion}
This paper proposed a novel multitask learning-based model, \textsf{AngryBERT}, which jointly learns hate speech detection with emotion classification and target identification as secondary relevant tasks.  We evaluated \textsf{AngryBERT} on three publicly available real-world datasets, and our extensive experiments have shown that \textsf{AngryBERT} outperforms the state-of-the-art single-task and multitask baselines in the hate speech detection tasks. We identify case studies to demonstrate that \textsf{AngryBERT} is able to detect hate speeches accurately and identify the target of the hate speech and the emotion expressed. For future works, we will explore developing better approaches to automatically select the optimal combination of co-training tasks for hate speech detection. We will also explore developing explainable hate speech detection methods that utilized the predictions of secondary tasks as supplementary information.

%
% ---- Bibliography ----
%
% BibTeX users should specify bibliography style 'splncs04'.
% References will then be sorted and formatted in the correct style.
%
\bibliographystyle{splncs04}
\bibliography{ref}
%
% \begin{thebibliography}{8}
% \bibitem{ref_article1}
% Author, F.: Article title. Journal \textbf{2}(5), 99--110 (2016)

% \bibitem{ref_lncs1}
% Author, F., Author, S.: Title of a proceedings paper. In: Editor,
% F., Editor, S. (eds.) CONFERENCE 2016, LNCS, vol. 9999, pp. 1--13.
% Springer, Heidelberg (2016). \doi{10.10007/1234567890}

% \bibitem{ref_book1}
% Author, F., Author, S., Author, T.: Book title. 2nd edn. Publisher,
% Location (1999)

% \bibitem{ref_proc1}
% Author, A.-B.: Contribution title. In: 9th International Proceedings
% on Proceedings, pp. 1--2. Publisher, Location (2010)

% \bibitem{ref_url1}
% LNCS Homepage, \url{http://www.springer.com/lncs}. Last accessed 4
% Oct 2017
% \end{thebibliography}
\end{document}